\documentclass[a4paper,fleqn]{cas-dc}

\usepackage[numbers]{natbib}
\usepackage{microtype}
\usepackage{graphicx}
\usepackage{amsthm}
\usepackage{amsmath}
\usepackage{hyperref}
\usepackage{footnote}

\usepackage{soul}

\def\tsc#1{\csdef{#1}{\textsc{\lowercase{#1}}\xspace}}
\tsc{WGM}
\tsc{QE}
\tsc{EP}
\tsc{PMS}
\tsc{BEC}
\tsc{DE}

\begin{document}
\renewcommand{\printorcid}{}
\let\WriteBookmarks\relax
\def\floatpagepagefraction{1}
\def\textpagefraction{.001}
\shorttitle{Deep DA with Ordinal Regression for Pain Assessment Using Weakly-Labeled Videos}
\shortauthors{Praveen et~al.}
\title [mode = title]{Deep Domain Adaptation with Ordinal Regression for Pain Assessment Using Weakly-Labeled Videos}                

\author[mymainaddress]{R Gnana Praveen} 
\cormark[1]
\ead{gnanapraveen.rajasekar.1@ens.etsmtl.ca}
\author[mymainaddress]{ Eric Granger}
\ead{Eric.Granger@etsmtl.ca}
\author[mymainaddress]{ Patrick Cardinal}
\ead{Patrick.Cardinal@etsmtl.ca}


\cortext[mycorrespondingauthor]{Corresponding author}
\address[mymainaddress]{Laboratoire d’imagerie, de vision et d’intelligence artificielle\\
École de technologie supérieure, Université du Québec\\
Montreal, Canada\\}




\begin{abstract}
Estimation of pain intensity from facial expressions captured in videos has an immense potential for health care applications. Given the challenges related to subjective variations of facial expressions, and to operational capture conditions, the accuracy of state-of-the-art deep learning (DL) models for recognizing facial expressions may decline. Domain adaptation (DA) has been widely explored to alleviate the problem of domain shifts that typically occur between video data captured across various source (laboratory) and target (operational) domains. Moreover, given the laborious task of collecting and annotating videos, and the subjective bias due to ambiguity among adjacent intensity levels, weakly-supervised learning (WSL) is gaining attention in such applications. State-of-the-art WSL models are typically formulated as regression problems, and do not leverage the ordinal relationship among pain intensity levels, nor the temporal coherence of multiple consecutive frames. This paper introduces a new DL model for weakly-supervised DA with ordinal regression (WSDA-OR) that can be adapted using target domain videos with coarse labels provided on a periodic basis. The WSDA-OR model  enforces ordinal relationships among the intensity levels assigned to target sequences, and associates multiple relevant frames to sequence-level labels (instead of a single frame). In particular, it learns discriminant and domain-invariant feature representations by integrating multiple instance learning with deep adversarial DA, where soft Gaussian labels are used to efficiently represent the weak ordinal sequence-level labels from the target domain. The proposed approach was validated using the RECOLA video dataset as fully-labeled source domain data, and UNBC-McMaster shoulder pain video dataset as weakly-labeled target domain data. We have also validated WSDA-OR on BIOVID and Fatigue (private) datasets for sequence level estimation. Experimental results indicate that our proposed approach can significantly improve performance over the state-of-the-art models, allowing to achieve a greater pain localization accuracy.  Code is available on GitHub link: \url{https://github.com/praveena2j/WSDAOR}.
\end{abstract}



\maketitle

\section{Introduction}

\begin{figure}[t!]
\centering
\includegraphics[width=0.45 \textwidth]{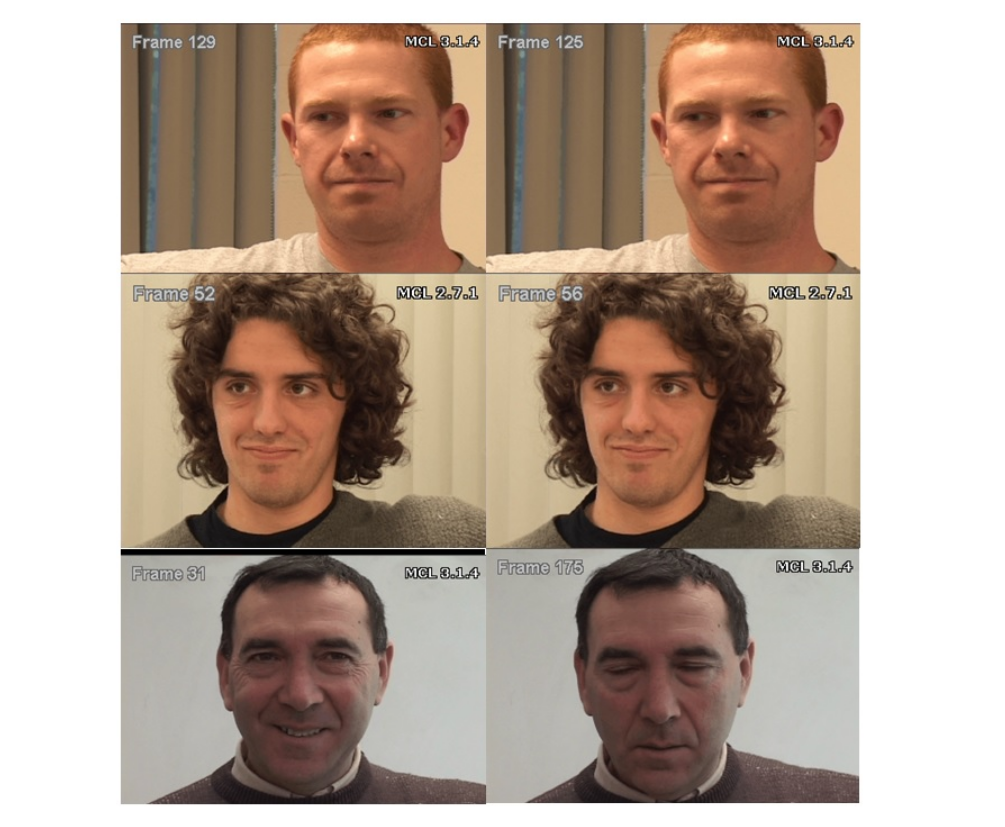}
\caption{ \textbf{Examples of video frames with (left) and without (right) shoulder pain from the UNBC-McMaster dataset \cite{5771462}.}}
\label{fig:pain}
\end{figure}

Pain is a highly disturbing sensation caused by injury, illness or mental distress. It is a primitive symptom for the malfunctioning of any system in our body \cite{pain:2001}. Pain is typically conveyed through a patient, or an observer on a linear scale from $0$ (no pain) to $10$ (severe pain). However, assessment provided by a patient or an observer may not be reliable since it is subjected to bias induced by the individual's perception of pain as shown in Figure \ref{fig:pain}. Automatic estimation of pain is useful for people who lack verbal communication such as infants, patients suffering from neurological disorder, those in intensive care unit (ICU) requiring assisted breathing, etc. Therefore, there is a growing demand for the development of automatic pain management system to ensure effective treatment and ongoing care. 

One of the primary channels through which pain can be effectively communicated are facial expressions. Over the years, there has been significant progress on automatic estimation of pain intensities based on facial expressions in videos \cite{8928510}. In the recent years, deep learning (DL) models have provided state-of-the-art performance in many visual recognition applications such as object detection, image classification, semantic segmentation, action recognition, etc \cite{8627998}. Compared to 2D-CNN models, 3D-CNNs are found to be efficient for encoding the spatiotemporal dynamics of facial expressions in videos \cite{8756568}. However, using DL models poses several challenges for real-world pain intensity estimation.  An important challenge is the subjective variability of facial expressions across different individuals, and operational capture conditions of videos. Indeed, the performance of DL models for facial expression recognition may decline significantly when there is a considerable domain shift between data distributions of videos captured in the source (lab setting) and target (operational) domains \cite{WANG2018135}. Domain adaptation (DA) has been widely used to address the problem of domain differences in various visual recognition applications \cite{WANG2018135}. In particular, unsupervised DA (UDA) is commonly used for applications related to facial analysis, such as smile detection, in order to learn robust domain-invariant CNN representations based on labeled source and unlabeled target domain data \cite{Sangineto, Xiaoqing,7550085}. The literature on UDA techniques focuses on learning discriminant domain invariant embeddings by optimizing an adversarial loss to encourage domain confusion, or a discrepancy loss between the two data distributions. Reconstruction-based approaches are another popular paradigm to learn the mapping between source and target images such that images captured in different domains have similar appearance. 

In contrast with the existing DA approaches for facial expression analysis, we explore the weakly-supervised DA (WSDA) case, were source data is fully labeled (at a frame level), and target data is weakly labeled. The authors have explored deep DA models to learn a common representation that diminishes the domain shift between source and target domains. A preliminary version of the approach proposed in this paper is presented in \cite{Praveen}, where deep DA is explored for weakly-supervised pain localization in videos. In the present work, our approach is improved considerably by leveraging the ordinal relationship among intensity levels, and temporal coherence of multiple consecutive frames. We also provided a more detailed formulation and experimental validation of our method.

Performing DA for pain intensity level estimation from videos of faces is a  challenging problem, in particular when the reference video data is provided with a limited amount of annotations. Most of the existing DL models for pain intensity level estimation have been explored in the fully supervised setting,  using frame-level labels \cite{Tavakolian2018DeepSR, ZhouHSZ16}. However, annotating the pain intensity levels for large-scale datasets involves a costly and time-consuming process with domain experts. Moreover, manual annotation process is vulnerable to subjective bias, resulting in ambiguous labels.

Recently, weakly supervised learning (WSL) has been gaining attention for its potential to train machine learning (ML) models using data with a limited amount of annotations \cite{WSL}. Based on the availability of labels, WSL scenarios can be classified according to three categories: incomplete, inexact and inaccurate supervision \cite{WSL}. Incomplete supervision refers to the scenario where annotations are only provided for a subset of the training dataset. In scenarios involving inexact supervision, annotations are provided for the entire dataset, but at global or coarse level compared to ones provided in fully supervised scenario. Inaccurate supervision deals with scenarios where annotations are noisy and ambiguous. Praveen et al. \cite{r2021weakly} provided a comprehensive review of WSL based approaches for facial behavior analysis. In the context of pain assessment in videos, inexact supervision is a relevant scenario since it assumes that pain intensities are annotated on periodic basis or for entire videos (sequence-level), rather than at the frame level. In particular, multiple instance learning (MIL) methods has been widely used in applications such as object recognition \cite{Miao}, text categorization \cite{Andrews:2002:}, and context-based image retrieval \cite{CBIRMIL}, to train ML models using data with coarse annotations. Therefore, we have formulated the problem of pain intensity level estimation from faced captured in videos in the framework of MIL, where sequences are considered to be bags and frames as instances. Most MIL methods proposed in the literature for pain intensity estimation rely on handcrafted features and conventional ML approaches \cite{SIKKA2014659, 7163116, 8347018}, due in part to the limited availability of training data with sequence-level annotations. In this paper, we investigate deep WSDA models of pain intensity levels using sequence-level labels. 

Pain level assessment can be formulated as classification or regression problem. In classification, pain estimation is often formulated as a binary problem, i.e., pain/no-pain, whereas regression allows to predict a wider range of pain intensities. Recently, approaches based on the regression formulation have gained much attention in the literature because they provide more accurate localization of pain intensities \cite{Tavakolian2018DeepSR}, \cite{ZhouHSZ16}. Regression-based approaches can in-turn be classified into ordinal and continuous regression. Although continuous regression predicts a wider range of pain intensities, discrete pain intensity levels are often preferred in practical applications to ease video analysis and annotation. Ordinal relations among pain intensity levels which conveys a rich source of information, yet very few approaches have explored the ordinal relationship among pain intensity levels for automatic pain intensity estimation in videos \cite{8347018}. The problem of ordinal regression has been widely explored for various applications, such as age estimation \cite{7780901}, and image ranking \cite{10.1145/2072298.2072023}. However, the ordinal regression framework is less explored for pain intensity estimation in videos annotated at the sequence-level. This paper introduces a new deep WSDA model with ordinal regression (WSDA-OR). It learns discriminant and domain-invariant feature representations by integrating MIL with adversarial DA, where soft Gaussian labels are used to represent the weak ordinal sequence-levels from target videos. 

In most of conventional MIL approaches for pain assessment, MIL pooling is performed using maximum operator, i.e., the sequence level label is associated to the frame corresponding to the maximum intensity level \cite{6755527,Sikka:2014}. However, the maximum operator only relies on a single frame, failing to capture the relevant information available in multiple adjacent frames. Maximilian et al. \cite{pmlr-v80-ilse18a} have shown that attention based MIL pooling can significantly improve predictive accuracy. Inspired by their approach, we introduce adaptive MIL pooling, that relies on multiple relevant frames of the sequence (bag). It allows associating all the relevant frames of the corresponding sequence to the sequence-level label, and can significantly improve the accuracy of pain assessment.
To the best of our knowledge, WSDA-OR is the first model to efficiently capture the ordinal relationship among pain intensity levels through Gaussian representation, in the context of multiple instance regression (MIR).

The main contributions of this papers are: 
\begin{itemize}
    \item a DL model for pain assessment that can adapt to diverse capture conditions and individuals using weakly-labeled target videos; 
    \item Gaussian modeling through multiple instance regression (MIR) to efficiently capture the ordinal relationship among intensity levels;  
    \item an adaptive MIL pooling to associate all the relevant frames of the corresponding sequence to the sequence-level label;
    \item an extensive set of experiments validating that our proposed WSDA-OR can outperform  state-of-the-art models.
\end{itemize} 
The rest of this paper is organized as follows. Section \ref{sec:work} provides some background on models for pain intensity estimation, deep DA, ordinal regression, and MIL. Our proposed WSDA-OR model is described in Section \ref{sec:wsod}. Finally, Section~\ref{sec:res} presents the experimental methodology (datasets, protocols and performance metrics), and results for validation.


\section{Related Work}\label{sec:work}

\subsection{Deep Models for Pain Intensity Estimation:}
Though DL are explored extensively for fully supervised learning, it is still at rudimentary level to deal with weakly-labeled data. Jing et al \cite{ZhouHSZ16} proposed Recurrent Convolutional Neural Network (RCNN) using recurrent connections in the convolution layers to capture the temporal information without increasing the overload of parameters to avoid over fitting. In order to deal with the problem of limited data, Wang et al \cite{8296449} used a pretrained face recognition network for fine-tuning using a regularized regression loss. Rodrigue et al \cite{7849133} also used VGG Face pre-trained CNN network \cite{Parkhi15} for capturing the facial features and LSTM network is used to exploit the temporal relation between the frames. Compared to 2D CNN models, 3D CNNs are found to be gaining attention in efficiently capturing the temporal dynamics of the video sequences. Tavakolian et al \cite{Tavakolian2018DeepSR} propose a 3D-CNN based architecture using a stack of convolution modules with varying kernel depths for efficient dynamic spatiotemporal representation of faces in videos. A temporal pooling method to encode the spatio-temporal facial variations in video clips  based on two-stream model that performs late fusion of appearance and dynamic information \cite{9054375}. However all these approaches have been proposed in the setting of fully supervised learning, thereby requires frame-level labels. Inflated 3D-CNNs (I3D) have been employed for facial expression recognition, allowing to leverage pre-trained 2D-CNNs, 
yet benefit from the efficient modeling of temporal dynamics using 3D CNN models \cite{Ayral_2021_WACV, 8099985}. Inspired by these benefits and their performance, we have rely on I3D for modeling the spatiotemporal dynamics of pain expressions for adversarial DA with weakly-labelled target videos.

\subsection{Deep Domain Adaptation:}
Wang \textit{et al}. \cite{WANG2018135} provided a survey of the deep DA approaches, with applications in visual recognition. Deep DA can be primarily summarized to three categories : discrepancy based, adversarial based and reconstruction based DA. Discrepancy based approaches attempts to minimize the domain-shift by fine-tuning the deep model with the labeled or unlabeled target data. Adversarial based approaches deploys domain discriminators to classify whether a data sample is drawn from source or target domain to diminish the domain-shift. Finally, reconstruction based approaches tries to ensure feature invariance using data reconstruction of source or target samples to improve the performance of DA. Enver et al \cite{Sangineto} proposed a regression framework for personalized facial expression recognition, where classifiers are generated for the individuals of the source data rather than a generic model for the entire source data. Wang et al \cite{Xiaoqing} proposed unsupervised DA approach for small target dataset using Generative Adversarial Network (GAN), where GAN generated samples are used to fine-tune the model pretrained on the source dataset. Zhu et al \cite{7550085} explored unsupervised DA approach in the feature space, where the mismatch between the feature distributions of the source and target domains are minimized still retaining the discriminative information among the face images related to facial expressions. Behzad et al \cite{BOZORGTABAR2020107111} investigated the use of adversarial DA to transform the visual appearances of simulated faces to real face images without loosing the face details relevant to identity or expressions. By doing so, expression recognition models trained on labeled realistic face images with arbitrary head poses can be directly generalized on the unlabeled simulated images without the need for re-training. 

Contrary to the existing DA approaches for facial expression analysis, we have explored DA in the context of adapting source data with full labels to target data with coarse labels. Ganin et al \cite{Ganin:2015:UDA:3045118.3045244} proposed a novel approach of adversarial DA using deep models with partial or no target data labels using a simple gradient reversal layer. We have further extended their approach for the scenario of coarsely labeled target data for pain localization in videos.

\subsection{Ordinal Regression:}
Rui et al \cite{7780746} proposed a max-margin based ordinal support vector regression using ordinal relationship, which is flexible and generic in handling varying level of annotations. A linear model is learned by solving the optimization problem using Alternating Direction Method of Multipliers (ADMM) to predict the frame-level intensity of test image. Zhang et al \cite{zhangbilateral:2018} explored domain knowledge of Ordinal relevance, intensity smoothness and relevance smoothness based on the gradual evolving process of facial behavior. Yong et al \cite{Zhang_2018_CVPR} designed a CNN model for intensity estimation of Action Units(AUs) using annotations of only peak and valley frames, where the parameters of CNN are learned by encoding domain knowledge of facial symmetry, temporal intensity ordering, relative appearance similarity and contrastive appearance difference. All of the above mentioned approaches did not efficiently capture the ordinal relationship and are proposed for expression or action unit intensity estimation but not for pain intensity estimation.

\subsection{ Multiple Instance Learning:}
Though MIL has been widely explored for many computer vision applications, relatively fewer techniques have been proposed for dynamic pain intensity estimation. Sikka \textit{et al.} \cite{SIKKA2014659} developed automatic pain recognition system for pain localization in the framework of MIL, where video segments are represented as bags of multiple subsequences and MILBOOST \cite{MILBOOST} is used for instance-level pain detection. Chongliang et al \cite{7163116} further enhanced the approach by incorporating discriminative Hidden Markov Model (HMM) based instance level classifier in conjunction with MIL framework instead of MILBOOST to  efficiently  capture  the  temporal  dynamics. Chen et al \cite{Zhanli} proposed a novel two stage approach for pain detection by deploying a novel strategy to encode AU combinations using individual AU scores. 

However, all of these approaches have been proposed for pain detection. Ruiz et al. \cite{8347018} proposed multi-instance dynamic ordinal random fields (MI-DORF) for modeling temporal sequences of ordinal instances, where bags are defined as temporal sequences labeled as ordinal variables. The instance labels are obtained by incorporating high order cardinality potential relating bag and instance labels in the energy function. But they have not leveraged the superior performance of DL models. Yong et al. \cite{Zhang_2018_CVPR} designed a deep CNN based on weakly supervised learning for intensity estimation of Action Units(AUs) of facial expressions with limited annotations of AUs, where only the annotations of peak and valley frames of the AUs are considered. Despite the advancement of MIL for various applications in computer vision, not much work has been explored for the estimation of pain intensity levels using state-of-the-art DL models. Unlike the above mentioned approaches, our approach focus on pain intensity level estimation using DL models in conjunction with MIL framework for localization of pain intensity levels i.e., instance level prediction. We have further improved the pooling mechanism by introducing adaptive MIL pooling to efficiently leverage all the relevant frames in the sequence to associate with the sequence level label.

\section{Proposed Approach}\label{sec:wsod}

\begin{figure*}
\centering
\includegraphics[width=1.0\linewidth]{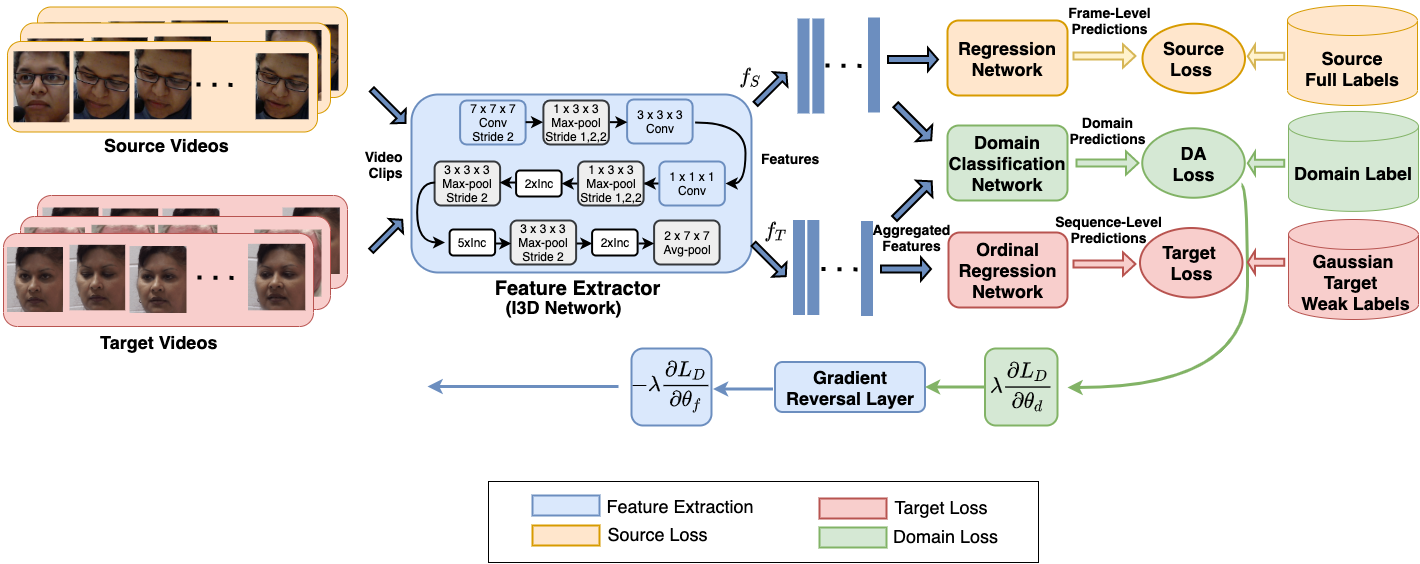}
\caption{\label{fig:WSDA-OR} \textbf{Overall Architecture of the proposed approach (WSDA-OR). Inc denotes Inception module \cite{7298594}. Different colors are used to discriminate data flow in different loss components. Best viewed in color}}
\end{figure*}

In this section, we elaborate the proposed approach in detail
\footnotemark
\footnotetext[1]{Code on GitHub link: \url{https://github.com/praveena2j/WSDAOR}.}. 
In the proposed framework, we have explored deep DA in the context of MIL for ordinal regression, where labels of intensity levels are provided for video sequences instead of individual frames. In order to efficiently model the ordinal relationship among the intensity levels, we have considered Gaussian modeling of the intensity levels (labels) instead of one-hot vectors. Unlike the conventional approaches of MIL \cite{Praveen}, \cite{8347018}, \cite{Sikka:2014}, where the sequence level label was associated with a single frame, we have exploited multiple frames, which are relevant to the sequence level label in order to enhance the performance of learning framework. Inspired by the performance of I3D model \cite{8099985} with adversarial learning \cite{Ganin:2015:UDA:3045118.3045244}, we have used the framework of adversarial based DA as it was shown to yield superior performance in the framework of DL models for videos \cite{Jamal2018DeepDA}. The overall block diagram of the proposed approach is shown in Fig \ref{fig:WSDA-OR}. 

Let $\mathbf{D} = \{ (\mathbf{X_1},\mathbf{Y_1}),(\mathbf{X_2},\mathbf{Y_2}),.......,(\mathbf{X_N},\mathbf{Y_N})\} $ represents the dataset of pain expressions of videos from source and target domains. $\mathbf{X_i}$ denotes a video sequence of the training data with a certain number of frames. In case of source domain, $\mathbf{Y_i}$ denotes a structured label vector with frame level annotations of the corresponding video sequence $\mathbf{X_i}$, whereas for target domain, $\mathbf{Y_i}$ represents an ordinal intensity value i.e., sequence level ordinal intensity value of the corresponding video sequence, which is given by 
\begin{equation}
\mathbf{Y_i}=\left\{\begin{array}{l}\{y_i^1,\;y_i^2,\;....y_i^{n_i}\}\;if\;\mathbf{X_i}\in\;source\;domain\\\;\;\;\;\;\;\;\;\;\;\;\;\;y_{i\;\;}\;\;\;\;\;\;\;\;if\;\mathbf{X_i}\in target\;domain\end{array}\right.
\end{equation}
where ${n_i}$ denotes the number of frames in the corresponding sequence $\mathbf{X_i}$. $N$ represents the number of training sequences. Specifically, $\mathbf{X_i} = \{ x_i^1,x_i^2,....x_i^{n_i}\} $ represents the temporal sequence of ${n_i}$ observations (frames) and $x_i^t$ denotes $t^{th}$ frame in $i^{th}$ sequence, where $t \in \{ 1,2,......,{n_i}\} $. 

The objective of the problem is to estimate a generic ordinal regression model $F:\mathbf{X} \to \mathbf{H}$ from the training data $\mathbf{D}$ in order to predict the pain intensity level of frames of unseen test sequences, where $\mathbf{X}$ denotes the video sequences of training data and $\mathbf{H}$ represents the hidden label space of frame-level annotations of the target domain. The estimated intensity levels  of the individual frames of the sequences in the target domain is predicted as structured output $\mathbf{H_i} \in \mathbf{H}$, where 
$\mathbf{H_i} = \{ h_i^1,h_i^2,.......,h_i^{n_i}\}  $ and each frame $x_i^t$ of the sequence is assigned a latent ordinal value ${h_i^t}$.

Let $S$ represents the source dataset, which is fully labeled videos and $T$ denotes the target dataset, which is weakly labeled videos. Let ${G_f}$  represents the feature mapping function, where the parameters of this mapping are denoted by ${\theta _f}$. Similarly, the feature vectors of source domain and target domain are mapped to the corresponding labels using ${G_l}$ and ${G_{wl}}$, whose parameters are denoted by ${\theta _l}$ and ${\theta _{wl}}$ respectively. Finally, the mapping of feature vector to the domain label is obtained by ${G_d}$ with parameters ${\theta _d}$.

\subsection{Gaussian Modeling of Ordinal Intensity Levels (GM):} 
\label{Gaussian labels}
Due to the ordinal nature of pain intensity levels, we have formulated pain intensity estimation as ordinal regression problem, which attempts to solve classification problem still retaining the ordinal relationship among the labels. Though ordinal regression can be formulated as a classification problem, it does not capture the relative ordering among the ordinal labels. For instance, if a particular sample has a pain intensity level of "4", a misclassification of "3" or "5" is more acceptable than a misclassification of "1". Although the objective is to predict the correct intensity level of "4", the system should, in the event of misclassification, logically predict an ordinal level as close as possible to the ground truth "4". 

\begin{figure}[htp]
\centering
\includegraphics[width=1.0\linewidth]{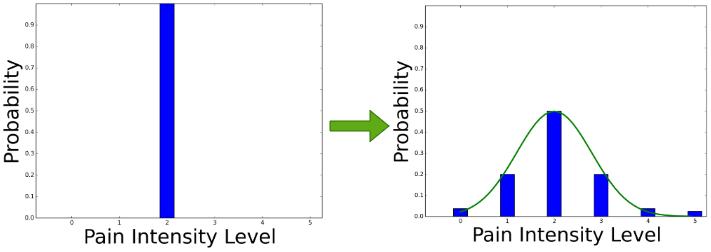}
\caption{\textbf{Gaussian Representation of weak ordinal labels}}
\label{fig:Gaussian Labels}
\end{figure}

In order to model the relative ordering among the ordinal labels, soft labels has been widely explored in the literature \cite{8953836, 10.1007/978-3-319-54187-7_14}. Gaussian distribution was found to be promising in modeling the ordinal relationships, where the probability (or contribution) of the nearby ordinal labels decreases exponentially as we move away from the ground truth on either sides of the mean in a symmetric manner. This approach has been widely explored in the ordinal regression literature
\cite{JMLR:v6:chu05a}, \cite{9009094}. Most of these approaches have imposed the constraints of Gaussian modeling on the predicted outputs. In this work, we propose a simple yet efficient approach of encoding the target labels as soft labels obtained from Gaussian distribution instead of one-hot vectors as shown in Fig \ref{fig:Gaussian Labels}. Specifically, the mean of the Gaussian model is considered as the corresponding ground truth label, and the variance controls the influence of neighboring ordinal levels. The intensity levels in close proximity to the corresponding label therefore have higher relevance compared to the intensity levels at far proximity. The soft Gaussian labels of the ordinal intensity levels are given by 
\begin{equation} 
q_i=e^{\textstyle\frac{-{(k-y_i)}^2}{2\sigma^2}}
\end{equation}
where $\sigma$ denotes Gaussian smoothing parameter (variance) of the Gaussian model and $k\in\{0,1,2,...,K-1\}$ and $K$ denotes the number of ordinal intensity levels.

The proposed approach of encoding the target labels using Gaussian distribution automatically learns the ordinal relationships without any explicit modification to the network architecture. Therefore, our method can also be used with any conventional classification networks with common categorical loss functions such as cross entropy. Additionally, deploying a soft Gaussian version of the target labels also helps in counterfeiting the problem of limited data with deep networks. We show empirically that these soft representations obtained from Gaussian distribution efficiently capture the ordinal relationship among the pain intensity levels, and significantly improve the performance of the system.

\subsection{Adaptive MIL Pooling (AMILP):} 
\label{Multiple Frames}
In the framework of MIL, choice of the pooling function plays a crucial role in associating the instance-level outputs to the bag label. Several pooling functions have been explored in the literature and a comparative study of various pooling techniques are discussed in \cite{8682847}. In the conventional setting of MIR \cite{6755527}, the sequence level label is associated with the frame corresponding to the highest intensity level in the framework of MIL \cite{Praveen, 8347018, Sikka:2014}. The relationship between the coarse bag-level label $\mathbf{Y_i}$ and latent instance level labels $\mathbf{H_i}$ is modeled by assigning the maximum value of predicted instance level outputs to the bag label, which is given by
\begin{equation}
{\mathbf{Y_i}} = \mathop {\max }\limits_h ({\mathbf{H_i}}){\rm{     }}~~~~~~\forall ({\mathbf{X_i}},{\mathbf{Y_i}}) \in {\mathbf{D}}
\end{equation}
If the label ${\mathbf{Y_i}}$ is $0$, then all the frames in the sequence ${\mathbf{X_i}}$ will be assigned $0$ i.e., neutral frame.

In case of pain intensity levels, the sequence level label is associated with the frame corresponding to the highest intensity level in the framework of MIR \cite{Praveen}, \cite{8347018, Sikka:2014}. The prediction of the weakly labeled sequence is given by
\begin{equation}
P(\mathbf{X_i}) = \mathop {\max }\limits_{j \in (1,..{n_i})} ({G_{wl}}({G_f}(x_i^j)))
\end{equation}
where $P(\mathbf{X_i})$ denotes the probabilities of the frame pertinent to maximum intensity level among all the frames of the sequence, $G_f$ and $G_wl$ represents the feature extraction and weak ordinal regression layers respectively.

However, maximum operator relies only on a single frame and does not efficiently exploit the information available in all the frames relevant to the sequence level label \cite{pmlr-v80-ilse18a}. In order to leverage the relevant information of multiple frames, several learnable pooling functions have been proposed with deep networks \cite{pmlr-v80-ilse18a, 8265256}. Unlike prior approaches, we have proposed a simple pooling function, which adaptively chooses the relevant frames without the need to learn any additional parameters. We further show that the proposed pooling mechanism has significantly improved the performance of the system.

The frames relevant to bag label (sequence level label) are selected based on the predicted instance-level outputs of the deep network. In case of pain intensity estimation, there could be many frames predicted as having the maximum pain intensity level, which are relevant to the sequence level label. The frames predicted as maximum intensity level are considered as frames relevant to sequence level label. Next, MIL pooling is performed by averaging the output responses of the selected relevant frames, whose outputs represents maximum intensity level within the sequence. Therefore, the frames that are irrelevant to the sequence -level (bag) label are discarded while the frames relevant to the sequence level label are retained and deployed in the pooling mechanism.

Typically, pain expressions are sparse in nature, where most of the frames in the sequence are neutral along with few pain expression frames relevant to sequence level labels. By deploying adaptive MIL pooling only on the frames pertinent to maximum predicted intensity levels, highly redundant neutral frames are discarded and the relevant multiple frames of higher intensity levels are effectively used in the training mechanism. For the sake of ordinal regression, the number of output units of the weak supervision layer ${G_{wl}}$ (last fully connected soft-max layer) of the ordinal regression module of target domain is equal to the number of intensity levels to be predicted. 
The bag level representation of the instance level outputs are obtained using adaptive MIL pooling, which is obtained by averaging of the outputs of selected frames (predicted as maximum intensity level) within the sequence, which is given by 
\begin{equation}
P(\mathbf{X_i}) = \frac{1}{{{N_{t_i}}}}\mathop {\sum }\limits_{j \in \max(1,..{n_i})} {G_{wl}}({G_f}(x_i^j))
\end{equation}
where $P(\mathbf{X_i})$ denotes the mean of the soft-max output responses of relevant frames predicted as maximum intensity levels and $N_{t_i}$ represents the number of relevant frames of the corresponding sequence predicted as maximum intensity level.

\subsection{WSDA-OR Training Mechanism:}

The deep network architecture consists of three major building blocks : feature mapping, label predictor and domain classifier. 
In the proposed architecture, the feature mapping layers share same weights between the source and target domains to ensure common feature space between source and target domains. It has been shown that the label prediction accuracy on the target domain will be same as that of the source domain by ensuring the similarity of distributions between source and target domains \cite{SHIMODAIRA2000227}. Next, adversarial mechanism is deployed between the domain discriminator ${G_d}$, which is learned to discriminate the source and target domain samples, and feature extractor ${G_f}$, which is trained simultaneously to minimize the domain discrepancy between source and target domains. At the time of training, label prediction loss is minimized on the source domain by optimizing the parameters of ${G_f}$ and ${G_l}$ in order to learn the feature mapping given the labels, while simultaneously ensuring the features to be domain-invariant. This is achieved by maximizing the loss of the domain classifier to minimize the discrepancy between the source and target domains while the parameters of ${G_d}$ are learned by minimizing the loss of domain classifier to discriminate source and target domains. The label prediction loss (${L_S}$) for the source domain is defined by 
\begin{equation}
{L_S} = \frac{1}{{{N_s}}}\sum\limits_{\mathop {i = 1}\limits_{{d_i} = 0} }^{N_s} {\sum\limits_{j = 1}^{{n_i}} {(({G_l}({G_f}(x_i^j))-y_i^j))^2} } 
\end{equation}
where ${{d_i} = 0}$ represents the source domain, ${N_s}$ denotes the number of video sequences in the source domain and ${n_i}$ denotes the number of frames in the corresponding video sequence. In addition to source labels, the weak labels of target domain is also used in the feature learning mechanism where the parameters of ${G_{wl}}$ are optimized by
minimizing the prediction loss pertinent to weak labels of the target data. The weak sequence level labels (ordinal intensity levels) of the target domain are encoded to soft Gaussian representations as mentioned in \ref{Gaussian labels} instead of one-hot vectors in order to efficiently capture the ordinal relationship as well as to counterfeit the problem of limited data. 

Contrary to MIL based approaches for pain intensity estimation, which relies on the single frame with maximum intensity level \cite{Praveen}, \cite{Sikka:2014}, we have explored multiple frames with maximum predicted intensity levels to associate with the weak sequence level label, thereby improving the training mechanism due to the deployment of multiple relevant frames as described in \ref{Multiple Frames}. The prediction loss associated with the weak supervision of the target domain is estimated as cross entropy (CE) loss between the soft gaussian labels of target domain $\mathbf{Y_i}$ and predicted response $P(\mathbf{X_i})$, which is given by    

\begin{equation}
{L_T} = -\frac{1}{{{N_T}}}\sum\limits_{\mathop {i = 1}\limits_{{d_i} = 1}}^{{N_T}} { \mathop (\mathbf{Y_i}.\log\left(P\left(\mathbf{X_i}\right)\right))} 
\end{equation}
where $\mathbf{Y_i}$ denotes the Gaussian representation vector of the ordinal level of the video sequence in the target domain, $P(\mathbf{X_i})$ denotes the vector of predicted intensity level of $\mathbf{X_i}$ in the target domain, $(.)$ denotes the dot product function and ${N_T}$ represents the number of video sequence in the target domain. 

Since domain classification is a typical binary classification problem, we have used logistic regression in order to diminish the domain differences between source and target domains, where the logistic loss function is given by 
\begin{multline}
{L_d} = \frac{1}{{{N_s} + {N_T}}}\sum\limits_{\mathop {i = 1}\limits_{{d_i} = 0,1} }^{{N_s} + {N_T}} {\sum\limits_{j = 1}^{{n_i}} [-{d_i^j\log\left({G_d}({G_f}(x_i^j))\right)}} \\
{- (1-d_i^j)\log\left(1-{G_d}({G_f}(x_i^j))\right)]}
\end{multline}
where  $d_i^j$ denotes the domain label of the ${j^{th}}$ frame of the ${i^{th}}$ video sequence.

The overall loss of the deep network architecture is given by
\begin{equation}
L = {L_S} + {L_T} - \lambda {L_d}
\end{equation}
where $\lambda$ is the trade-off parameter between the objectives of label prediction loss and domain prediction loss and the parameters of ${\theta _l}$, ${\theta _{wl}}$, ${\theta _f}$ and ${\theta _d}$ are jointly optimized using Stochastic Gradient Descent (SGD).

At the end of the training, the parameters of ${\theta _l}$, ${\theta _{wl}}$, ${\theta _f}$ and ${\theta _d}$ are expected to give a saddle point for the overall loss function as given by :
\begin{equation}
{{\hat \theta }_f},{{\hat \theta }_l},{{\hat \theta }_{wl}} = \mathop {\arg \min }\limits_{{\theta _f},{\theta _l},{\theta _{wl}}} L({\theta _f},{\theta _l},{\theta _{wl}},{{\hat \theta }_d})    
\end{equation}
\begin{equation}
{{\hat \theta }_d} = \mathop {\arg \max }\limits_{{\theta _d}} L({{\hat \theta }_f},{{\hat \theta }_l},{{\hat \theta }_{wl}},{\theta _d})    
\end{equation}
At the saddle point, the feature mapping parameters ${\theta _f}$ minimize the label prediction loss to ensure discriminative features and maximizes the domain classification loss to constrain the features to be domain-invariant. In order to backpropagate through the negative term in our loss function, a special gradient reversal layer (GRL) is deployed in our SGD optimization framework, which is elaborated in detail in \cite{Ganin:2015:UDA:3045118.3045244}. The value of lambda is modified over successive epochs, such that the supervised prediction loss dominates at
the early epochs of training. Further details on training mechanism can be found in \cite{Ganin:2015:UDA:3045118.3045244}.

\section{Results and Discussion}\label{sec:res}

\subsection{Experimental Setup:} 

The proposed approach has been evaluated on UNBC-McMaster dataset \cite{5771462}, which is widely used for pain intensity level estimation in the context of MIL. Due to the availability of state-of-the-art results of UNBC pain dataset in the context of MIL, we have primarily validated the proposed approach on UNBC pain dataset. The dataset consists of 200 videos of pain expressions captured from 25 individuals, resulting in 47,398 frames of size 320x240. Each video sequence is annotated using PSPI score at frame-level on a range of 16 discrete pain intensity levels (0-15). Due to the sparse nature of pain expressions and high level imbalance among various intensity levels, we followed the widely adapted quantization strategy i.e., the pain levels are quantized to 5 ordinal levels as: 0(0), 1(1), 2(2), 3(3), 4-5(4), 6-15(5). In our experiments, we followed the same experimental protocol as that of \cite{Praveen} in order to have a fair comparison with state-of-the-art results, where Leave-One-Subject-Out (LOSO) cross-validation strategy is deployed i.e., 15 subjects have been used for training, 9 subjects for validation and 1 for testing in each cycle.      

Due to the availability of labels for every frame, RECOLA \cite{6553805} dataset is used as source domain, where each video sequence is obtained for a duration of 5 minutes and annotated with an intensity value between -1 to +1 for every 40 msec (same as frame rate of 25fps) i.e., all the frames are annotated. The video sequences of UNBC (target) and RECOLA (source) datasets are converted to sub-sequences (bags) of 64 frames (instances) with a stride of 8 to generate more samples for the learning framework, resulting in 10496 sub-sequences for RECOLA and 2890 sub-sequences for UNBC dataset. In order to incorporate the setting of weakly supervised learning in target domain (weakly labeled videos), only coarse labels of the sub-sequences of UNBC dataset are considered i.e., the maximum intensity level within a sub-sequence is assigned as coarse annotation to formulate the problem of MIL for ordinal regression \cite{6755527}. In order to use the Gaussian representation of the weak ordinal labels, the variance is considered to be $0.3$.   

The faces are detected, normalized and cropped using MTCNN \cite{7553523} and resized to 224 x 224. In our experiments, I3D architecture is used, where inception v-1 architecture is used as base model, which is inflated from 2D pre-trained model on ImageNet to 3D CNN for videos of pain expressions. We have used Stochastic Gradient Descent (SGD) as our optimization technique for training the model with a momentum of 0.9, weight decay of 1e5. The initial learning rate is set to 0.001 and annealed according to a schedule pre-determined on the cross validation set for every 5 epochs after 20 epochs. Due to the difference between the number of samples (sub-sequences) between source and target domain, batch size of 4 is used for source domain and 2 for target domain. Due to the huge imbalance among various intensity levels of the samples (sub-sequences) of the target domain, weighted random sampling is deployed for loading the data to counterfeit the problem of level imbalance. An early stopping strategy is used for model selection to avoid over-fitting.

\subsection{Evaluation Measures:}
Given the ordinal nature of pain intensity levels, the performance of the proposed approach is measured in terms of Pearson Correlation Coefficient (PCC), Intra class correlation (ICC(3,1)) and Mean-Absolute-Error (MAE). In most of the existing literature in MIL, the results are often reported for bag-level predictions. However, we have focused on instance-level prediction i.e., frame-level prediction of ordinal pain intensity levels for accurate localization of pain intensity levels in videos. PCC is invariant to linear transformations and efficiently capture the correlation between predictions and ground truth, which may differ in scale and location. The PCC measure between predictions ($h_i$) and ground truth values ($y_i$) of a sequence $i$ is given by
\begin{equation}
    PCC(y_i,h_i)=\frac{n_i\sum(y_i\ast h_i)-{\displaystyle\sum_{}}y_i{\displaystyle\sum_{}}h_i}{\sqrt{\lbrack n_i{\displaystyle\sum_{}}y_i^2-{\displaystyle\underset{}{(\sum}}y_i)^2\rbrack\lbrack n_i\sum_{}h_i^2-\underset{}{(\sum}h_i)^2\rbrack}}
\end{equation}
where $n_i$ represents the number of frames in the sequence. 
Though PCC measure captures the correlation between the two variables, it fails to capture the exact similarity measure i.e., absolute agreement between ground truth and the predicted intensity levels. 
Therefore, we have used ICC(3,1) \cite{ICC}, which is widely used to accurately measure the degree of correlation as it takes into account differences in scale and location. The ICC measure between the predictions ($h_i$) and ground truth ($y_i$) is computed using Between Mean Squares (BMS) and Error Mean Squares (EMS), as given by 
\begin{equation}
    ICC(y_i,h_i)=\frac{BMS_i\;-\;EMS_i}{BMS_i\;+\;EMS_i}
\end{equation}
where $BMS_i$ of sequence $i$ is given by 
\begin{equation}
    BMS(y_i,h_i)=\frac{n_i{\displaystyle\sum_{}}{(y_i+h_i)}^2-{({\displaystyle\sum_{}}y_i+{\displaystyle\sum_{}}h_i)}^2}{2n_i(n_i-1)}
\end{equation}
and $EMS_i$ of sequence $i$ is given by 
\begin{equation}
    EMS(y_i,h_i)=\frac{2\sum y_i^2+2\sum h_i^2-\sum{(y_i+h_i)}^2}{2n_i}
\end{equation}

We have also further provided the performance measure of Mean-Absolute-Error (MAE), which is widely used for continuous regression applications and accurately captures the error between the two measurements of predictions ($h_i$) and ground truth ($y_i$), which is given by 
\begin{equation}
    MAE(y_i,h_i)=\frac{\sum\left|y_i-h_i\right|}{n_i}
\end{equation}


\begin{table}
\scriptsize
\renewcommand{\arraystretch}{1.2}
    \centering
\begin{tabular}{|l||c|c|c|c|c|c|c|c|c|c|} 
	\hline
	 \textbf{Training Scenario}  & \multicolumn{3}{|c|}{\textbf{Frame-level}}  \\
\cline{2-4}
	 & \textbf{PCC} $\uparrow$ & \textbf{ICC} $\uparrow$ &\textbf{MAE} $\downarrow$  \\
	\hline 	\hline
    Supervised (source data only) & 0.323 & 0.272 & 0.976 \\
	\hline
	Supervised (target data only) & 0.441 & 0.377  & 0.660 \\
	\hline
	Supervised (source $\cup$ target) & 0.570 & 0.448 & 0.539  \\
	\hline 	\hline
	Unsupervised DA     & 0.468 & 0.198 & 0.782 \\
	\hline
	Transfer learning with weak labels   & 0.614 & 0.384 & 0.618 \\
	\hline
	 Supervised DA & 0.750    &  0.724    & 0.440   \\
	\hline 
\end{tabular}
  \caption{ \textbf{PCC, ICC and MAE performance of proposed approach under various baseline scenarios.}}
    \label{results of various scenarios}
\end{table}

\subsection{Results with Baseline Training Models:}
To analyze the impact of DA and availability of annotations of source and target domains, the performance of the proposed approach has been evaluated by conducting a series of experiments with various baseline models, where I3D training models are generated by varying the data ranging from using only source data with full labels to the entire dataset of source and target domains with full labels as shown in Table \ref{results of various scenarios}. In all these experiments, the performance of the training model has been validated on the test data of the target domain. Initially, we have considered only the source domain with full labels without target domain and generated the training model. Due to the domain differences between train data (source) and test data (target), the generated training model exhibits poor performance. Next, we consider only the target data with full labels without source data and generate the training model, which shows improvement in performance as both train data and test data comes from the same domain (target). Subsequently, we use both source data and target data with full labels without DA 
and found that the performance was further improved as training data spans wide range of variation in source and target domains.    

\begin{figure}[htp]
\centering
\includegraphics[width=1.0\linewidth]{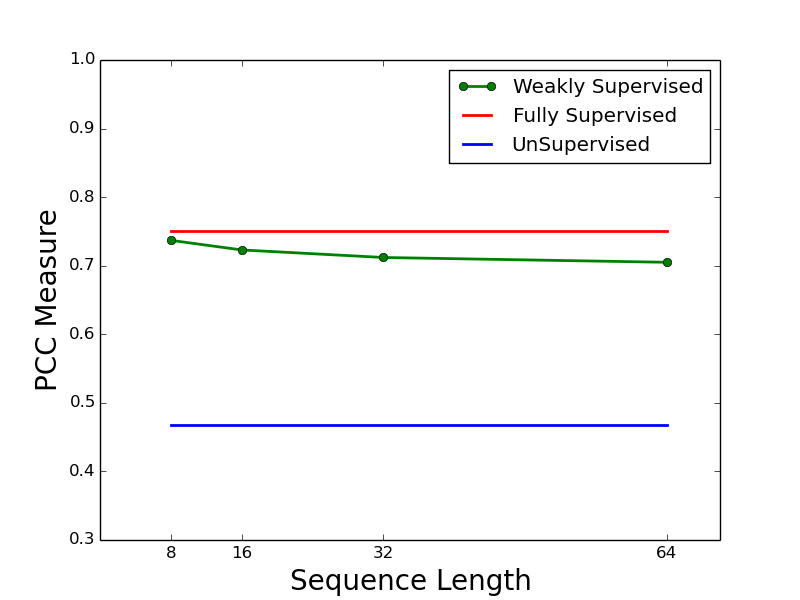}
\caption{\textbf{PCC accuracy of I3D model trained with deep WSDA-OR levels with decreasing level of weak supervision on target videos.}}
\label{fig:varyong sequence lengths}
\end{figure}

Now we conduct another series of experiments with DA, where the training data is obtained from source data with full labels and target data with varying levels of supervision. By considering the full labels of source domain, level of supervision of target domain are gradually reduced by decreasing the frequency of annotations i.e., labels are provided by increasing the duration of sequence lengths. Specifically, we have conducted experiments for sequence lengths of $8$,$16$,$32$ and $64$. In addition to varying sequence lengths, we have also conducted experiments of DA with no supervision of target domain, which acts as lower bound and full supervision of target domain, which acts as upper bound. As we gradually reduce the amount of labels of the target domain, we can observe that the performance of our approach gradually drops as shown in Fig \ref{fig:varyong sequence lengths}. However, our approach still performs at par with full supervision as there is only minimal decline, which is attributed to the DA as we are leveraging source data to adapt to target domain using adversarial DA. We have further evaluated the proposed approach with transfer learning, where the training model is first obtained only with source domain, and then find-tuned with the weak labels of target domain. Since transfer learning does not try to diminish the domain differences and rely on the size of pretrained dataset, it shows lower performance compared to the proposed approach.

\subsection{Ablation Study}

We have further 
analyzed the contribution of individual modules of the proposed approach : Gaussian modeling of ordinal levels (GM) and Adaptive MIL pooling (AMILP) as shown in Table \ref{results of ablation study}. First, we have generated the training model with a baseline version without GM and AMILP i,e., we have used max operator for MIL pooling and conventional label smoothing \cite{7780677}. The performance of the baseline training model is low as the conventional max-pooling operation fails to leverage the significant information in the nearby relevant frames and traditional label smoothing is not able to efficiently capture the ordinal relationships. Next, we have deployed AMILP of relevant frames without using GM of ordinal levels. This shows that AMILP efficiently leverages the information in the nearby relevant frames, thereby shows significant improvement over the conventional MIL pooling. In order to validate the contribution of GM of ordinal labels, we have further generated the training model only with the GM of ordinal labels over the baseline version. Since GM effectively captures the ordinal relationship among the pain intensity levels, the performance of the approach has significantly improved over conventional label smoothing. Finally, we have enforced GM of ordinal levels along with AMILP of relevant frames. By combining both the modules, there was drastic improvement in the performance of the approach as it effectively leverages all the relevant frames for MIL pooling and captures the ordinal relationship among the pain intensity levels.  

\begin{table}[htp]
\scriptsize
\renewcommand{\arraystretch}{1.2}
    \centering
\begin{tabular}{|l||c|c|c|c|c|c|c|c|c|c|} 
	\hline
	 \textbf{Training Scenario}  & \multicolumn{3}{|c|}{\textbf{Frame-level}}  \\
\cline{2-4}
	 \textbf{\ \ for WSDA-OR} & \textbf{PCC} $\uparrow$ & \textbf{ICC} $\uparrow$ &\textbf{MAE} $\downarrow$  \\
	\hline 	\hline
	Baseline & 0.511 & 0.498 & 0.632 \\
	\hline
	Baseline + AMILP & 0.627 & 0.597 & 0.740 \\
	\hline
	Baseline + GM & 0.598 & 0.599 & 0.617 \\
	\hline
	Baseline + GM + AMILP & 0.705 & 0.696 & 0.530 \\
	\hline

\end{tabular}
  \caption{ \textbf{PCC, ICC and MAE performance of proposed with ablation study of individual modules.}}
    \label{results of ablation study}
\end{table}

\begin{table*}
\scriptsize
\renewcommand{\arraystretch}{1.4}
    \centering
    \hspace{-12.9mm}
\begin{tabular}{|l|c||c|c|c|c|c|c|c|c|c|} 
	\hline
	 \textbf{Method} & \textbf{Type of}  & \multicolumn{3}{|c|}{\textbf{Frame-level}} & \multicolumn{3}{|c|}{\textbf{Sequence-level}}  \\
\cline{3-8}
	 & \textbf{Supervision} & \textbf{PCC} $\uparrow$ & \textbf{ICC} $\uparrow$ & \textbf{MAE} $\downarrow$   & \textbf{PCC} $\uparrow$ & \textbf{ICC} $\uparrow$ & \textbf{MAE} $\downarrow$  \\
	 \hline
	\hline
    MIR \cite{6755527} & Weak & 0.350 & 0.240 & 0.840  & 0.63 & 0.630 & 0.940   \\
	\hline
	 MILBOOST \cite{SIKKA2014659} & Weak & 0.280 & 0.110 & 1.770  & 0.380 & 0.380 & 1.700 \\
	\hline
	MI-DORF \cite{8347018} & Weak & 0.400 & 0.460 & 0.190  & 0.670 & 0.660 & 0.800 \\
	\hline
	WSDA \cite{Praveen} & Weak & 0.630 & 0.567  & 0.714  & 0.828 & 0.762 & 0.647  \\
	\hline 
	WSDA-OR (ours) & Weak & \textbf{0.705} & \textbf{0.696}  & \textbf{0.530}  & \textbf{0.745} & \textbf{0.750} & \textbf{0.443} \\
	\hline
	\hline
	BORMIR \cite{zhangbilateral:2018} & Semi & 0.605 & 0.531 & 0.821  & - & - & - \\
	\hline
	\hline
	LSTM \cite{7849133} & Full & 0.780 & - & 0.500  & - & - & -\\
	\hline
	SCN \cite{Tavakolian2018DeepSR} & Full & 0.920   & 0.750 & 0.320 (MSE) & - & - & - \\
	\hline
\end{tabular}
  \caption{ \textbf{Performance of the proposed WSDA-OR approach with state-of-the-art in terms of PCC, ICC, and MAE.}}
    \label{Comparison with state-of-the-art}
\end{table*}

\subsection{Comparison with State-of-Art Methods:}

\begin{figure*}
\centering
\includegraphics[width=0.95\linewidth]{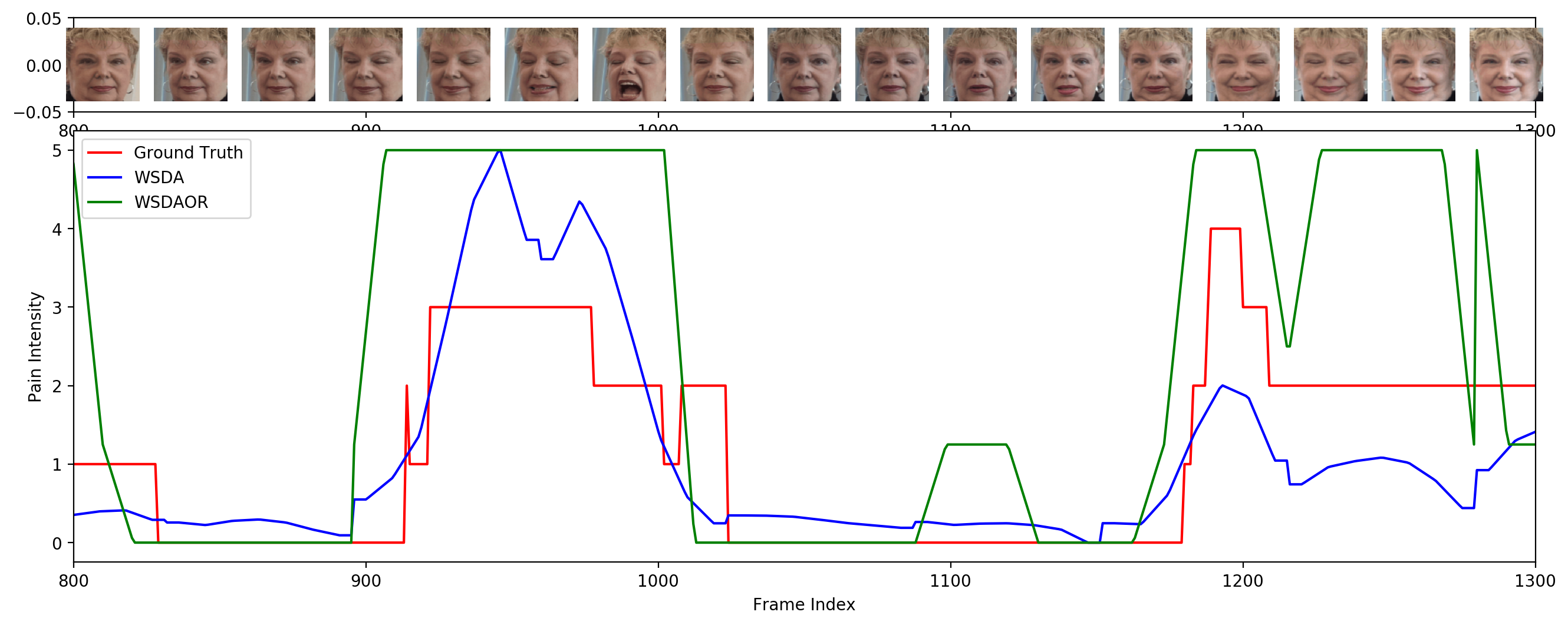}
\includegraphics[width=0.95\linewidth]{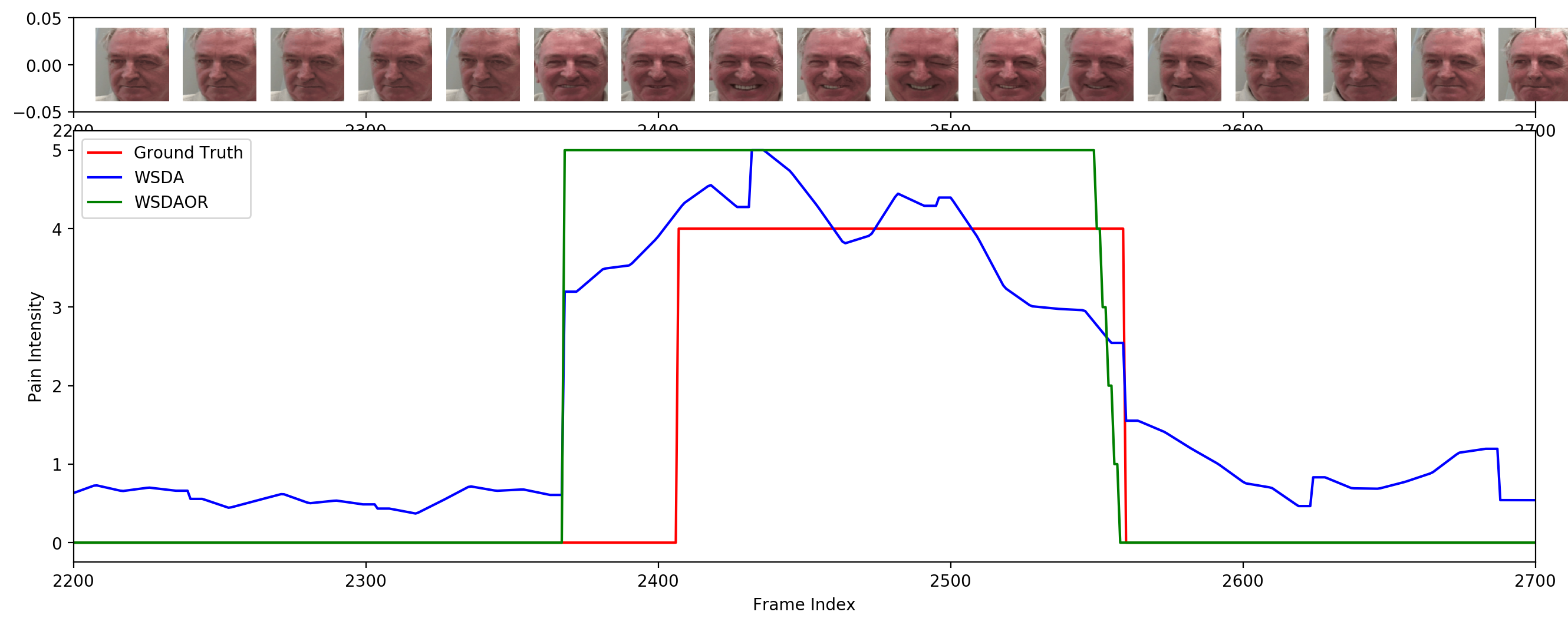}
\caption{\textbf{Visualization of pain localization on two different subjects in UNBC dataset. From top to bottom: Scenario with multiple peaks of pain expressions, our deep WSDA-OR localizes pain better than ground truth and WSDA \cite{Praveen}. Scenario where ground truth (GT) shows no pain, but our deep WSDA-OR approach correctly localizes pain better than WSDA \cite{Praveen}.}}
\label{pain loc vis}
\end{figure*}

In most of the existing state-of-the-art approaches for weakly supervised learning, classical ML approaches have been explored due to the problem of limited data with limited annotations. However, we have used deep model (I3D) along with source data to compensate the problem of limited data with limited annotations using DA. Our work is closely related to that of MI-DORF \cite{8347018}, which uses graph based models to capture the temporal dynamics of the frames and ordinal relationship of labels. We have further compared the proposed approach with our previous work \cite{Praveen} and shown that exploring the ordinal relationships and relevant frames in adaptive MIL pooling significantly improves the performance of the system over conventional regression framework. The estimation of frame level predictions of the proposed approach shows significant improvement, while sequence level estimation performs at par with that of our previous work \cite{Praveen}. This is due to the fact that the sequence level performance in the proposed approach is obtained from the model trained for frame-level predictions whereas the sequence level performance of our previous work \cite{Praveen} is reported using the model trained for sequence level predictions instead of frame level predictions. We have also provided visualization of some of our results for pain localization i.e., frame-level prediction for two subjects. Due to the efficient modeling of the ordinal labels and adaptive MIL pooling, the proposed approach accurately localizes pain better than \cite{Praveen} even though it was not captured in ground truth as shown in Figure \ref{pain loc vis}.

Since the problem of pain intensity estimation in the framework of ordinal regression with MIL is still at rudimentary level, we have also compared our approach with that of \cite{SIKKA2014659}, which was proposed for classification of pain events at sequence level. Unlike the existing classification approaches based on MIL, which estimates sequence level labels, we have estimated the ordinal intensity level of the individual frames using weak supervision of sequence level labels in target domain. Compared to the classification based approaches \cite{7163116}, \cite{SIKKA2014659}, regression based approaches \cite{8347018} are found to show superior performance due to the fact that intensity level estimation is closely related to the problem of regression. However, failing to exploit the discrete nature of labels shows poor performance in accurately tracking the pain intensity levels as that of the ground truth as shown in Figure \ref{pain loc vis}. Therefore, we have exploited the ordinal nature of the labels in our formulation, which significantly improves the performance of the approach and effectively tracks the pain intensity levels as reflected in the performance evaluation metrics as shown in Table \ref{Comparison with state-of-the-art}.

The proposed approach shows superior performance in terms of PCC compared to ICC and MAE, thereby effectively tracks the pain intensity levels of the individual frames as shown in Figure \ref{pain loc vis}. Since ICC is more reliable than PCC for sequence-level estimation as it efficiently captures the intra-class correlation, the proposed approach exhibits higher performance of ICC for sequence-level estimation compared to frame-level estimation. We have further compared the performance of the approach to that of state-of-the-art fully supervised \cite{Tavakolian2018DeepSR} as well as partially annotated scenarios \cite{zhangbilateral:2018}. The proposed approach performs better than Zhang et al \cite{zhangbilateral:2018} even without the requirement of any prior information pertinent to peak and valley frames.

\subsection{Results with Additional Datasets:}

The WSDA-OR model was also validated on Biovid and Fatigue (private) datasets. In our experiments, we have used Biovid PartA, which has 8700 videos of 87 subjects, which are labeled with pain levels of 0 to 4 at sub-sequence level. The Fatigue dataset is obtained from $18$ participants in Rehabilitation center, who are suffering from fatigue related issues. A total of 27 video sessions are captured from $18$ participants with a duration of 40 - 45 minutes and the videos are labeled at sequence level on a scale of 0 to 10 for every 10 to 15 minutes. However, due to the lack of frame-level labels and availability of state-of-the-art results for these datasets in the context of weak supervision, we have validated with other baseline models with transfer learning for sequence level estimation. With Biovid, 20 subjects are randomly selected for testing out of 87 subjects, whereas with Fatigue, on each trial we alternate testing on one subject out of 18 subjects. 

We have conducted experiments on three scenarios and the results are shown in Table \ref{Biovid}. In the first scenario, the model was trained using the Recola dataset as source domain data. Since the model is trained only on source domain, the model shows poor performance on the test (target) data since there is significant shift between the source and target domains. In the second case, the model trained using Recola as source domain data is further fine-tuned on Biovid as target domain data. This model shows improvement when compared to the first case, where the model is trained only on source data. Finally, we train a model using our WSDA-OR proposed approach, which shows significant improvement compared to the previous scenarios since it minimizes the domain differences and leverages the variability of both the domains to improve the generalization capability. 
   
\begin{table}
\scriptsize
\renewcommand{\arraystretch}{1.2}
    \centering
\begin{tabular}{|l||c|c|c|c|c|c|c|c|c|c|} 
	\hline
	 \textbf{Training Scenario}  & \multicolumn{3}{|c|}{\textbf{Sequence-Level}}  \\
\cline{2-4}
	 & \textbf{PCC} $\uparrow$ & \textbf{ICC} $\uparrow$ &\textbf{MAE} $\downarrow$  \\
    \hline \hline
\multicolumn{4}{|c|}{\textbf{Biovid Dataset}}      \\ 
	\hline 	
    Supervised (source data only) & 0.026 & 0.003 & 1.424 \\
	\hline
	Transfer learning & 0.246 & 0.240  & 1.242 \\
	\hline
	DA (proposed approach) & 0.341 & 0.317 & 1.162  \\
	\hline \hline
	
	\multicolumn{4}{|c|}{\textbf{Fatigue (Private) Dataset}}      \\ 
	\hline 	
    Supervised (source data only) & 0.028 & 0.007 & 1.645 \\
	\hline
	DA (proposed approach) & 0.436 & 0.367 & 0.363  \\
	\hline
\end{tabular}
  \caption{ \textbf{PCC, ICC and MAE performance of proposed WSDA-OR approach under different scenarios.}}
    \label{Biovid}
\end{table}

\section{Conclusion}

In this work, we have proposed a generic DL framework of weakly-supervised DA with ordinal regression for pain level assessment in videos. In order to address the problem of variations across different operational conditions, we have explored deep DA to leverage the performance of deep models by overcoming the problem of limited representative training data. We have formulated the framework of deep DA in the context of limited annotations, where the source domain is assumed to have fully supervised labels and target domain is assumed to have weak sequence level labels. 
Contrary to the existing approaches for pain intensity estimation, which explored DL models for regression, we have shown that exploiting ordinal relationships among the intensity levels significantly improves the performance of the system to accurately track and localize the pain intensity levels in videos. The ordinal intensity levels are modeled using a Gaussian distribution, which efficiently captures the ordinal relationships among the intensity levels.

In the conventional MIR approach \cite{6755527}, the weak sequence-level label is associated only with the frame pertinent to the frame with the maximum intensity level. However, we have improved the performance of the system by deploying multiple frames relevant to the sequence level label instead of a single frame. We have conducted extensive set of experiments with various baseline models using various combinations of source and target dataset and further analyzed the performance of the proposed approach under varying levels of supervision for the target data. Finally, we have compared the performance of the proposed approach with the state-of-the-art approaches and shown that the proposed approach significantly outperforms over the state-of-the-art approaches. 

\bibliographystyle{unsrt}
\bibliography{RevisedManuscript_Unmarked_edited}

\end{document}